\documentclass{article}

\usepackage{natbib}
\usepackage{arxiv}

\usepackage[utf8]{inputenc} 
\usepackage[T1]{fontenc}    
\usepackage{url}            
\usepackage{booktabs}       
\usepackage{amsfonts}       
\usepackage{nicefrac}       
\usepackage{microtype}      
\usepackage{graphicx}

\usepackage{hyperref}       
\usepackage{doi}
\usepackage{amsmath}
\usepackage{amssymb}
\usepackage{algorithm}
\usepackage{algorithmic}
\usepackage{rotating, booktabs}
\usepackage{tikz, pgfplots}
\usepackage{tabularray}
\usepackage{array}

\title{Transformer-based assignment decision network for multiple object tracking}

\date{} 					

\author{ \href{https://orcid.org/0000-0002-7412-0529}{\includegraphics[scale=0.06]{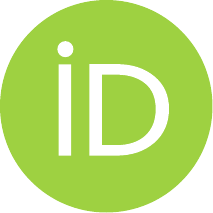}\hspace{1mm}Athena Psalta} \\
	Remote Sensing Laboratory\\
	National Technical University of Athens\\
	Iroon Polytechneiou 9, Athens 15780, Greece \\
	\texttt{psaltaath@central.ntua.gr} \\
	\And
	\href{https://orcid.org/0000-0003-2592-2127}{\includegraphics[scale=0.06]{orcid.pdf}\hspace{1mm}Vasileios Tsironis} \\
	Remote Sensing Laboratory\\
	National Technical University of Athens\\
	Iroon Polytechneiou 9, Athens 15780, Greece \\
	\texttt{tsironisbi@central.ntua.gr} \\
    \And
	\href{https://orcid.org/0000-0001-8730-6245}{\includegraphics[scale=0.06]{orcid.pdf}\hspace{1mm}Konstantinos Karantzalos} \\
	Remote Sensing Laboratory\\
	National Technical University of Athens\\
	Iroon Polytechneiou 9, Athens 15780, Greece \\
	\texttt{karank@central.ntua.gr} \\
}

\renewcommand{\headeright}{Preprint}
\renewcommand{\undertitle}{Preprint}
\renewcommand{\shorttitle}{Transformer-based assignment decision network for MOT}

\hypersetup{
pdftitle={A template for the arxiv style},
pdfsubject={q-bio.NC, q-bio.QM},
pdfauthor={David S.~Hippocampus, Elias D.~Striatum},
pdfkeywords={First keyword, Second keyword, More},
}

\begin{document}
\maketitle

\begin{abstract}
Data association is a crucial component for any multiple object tracking (MOT) method that follows the tracking-by-detection paradigm. To generate complete trajectories such methods employ a data association process to establish assignments between detections and existing targets during each timestep. Recent data association approaches try to solve either a multi-dimensional linear assignment task or a network flow minimization problem or tackle it via multiple hypotheses tracking. However, during inference an optimization step that computes optimal assignments is required for every sequence frame inducing additional complexity to any given solution. To this end, in the context of this work we introduce Transformer-based Assignment Decision Network (TADN) that tackles data association without the need of any explicit optimization during inference. In particular, TADN can directly infer assignment pairs between detections and active targets in a single forward pass of the network. We have integrated TADN in a rather simple MOT framework, designed a novel training strategy for efficient end-to-end training and demonstrated the high potential of our approach for online visual tracking-by-detection MOT on several popular benchmarks, i.e. MOT17, MOT20 and UA-DETRAC. Our proposed approach demonstrates strong performance in most evaluation metrics despite its simple nature as a tracker lacking significant auxiliary components such as occlusion handling or re-identification. The implementation of our method is publicly available at \url{https://github.com/psaltaath/tadn-mot}
\end{abstract}


\section{Introduction}
Multiple object tracking (MOT) aims at identifying objects of interest in a video sequence and representing them as a set of trajectories through time. In recent years, MOT has drawn significant attention due to its critical role in a wide range of applications such as traffic surveillance (\cite{traffic-surveillance}), autonomous driving (\cite{autonomous-driving},  \cite{autonomous-driving-2}) and human behavior prediction (\cite{psaltatsir}).

Due to recent advancements in object detection, most modern trackers (\cite{track-by-det}, \cite{tracktor}) follow a tracking-by-detection pipeline where objects of interest localized by a detector are associated by a predefined metric in order to form existing trajectories. Missing detections, interaction between targets in a crowded scene and partial or full occlusions of targets are only few of the challenges that MOT algorithms struggle with and can often lead to trajectory fragmentation and low performance. Under an online context, tracking-by-detection approaches rely heavily on data association methods in order to compose target trajectories through comparing detection hypotheses only with existing trajectories within a video frame. 

In most existing methods, this data association process relies on computing a set of similarity scores between existing trajectories and detections. In general, a significant number of modern MOT algorithms rely heavily on data association methods such as network flow optimization like \cite{network-flow-1} or the Hungarian algorithm (\cite{kuhn1955hungarian}). However, in any case a final optimization step is always required in order to infer the optimal assignments between multiple detections and existing targets adding extra complexity to a typical MOT algorithm. 

In this work, we have designed and developed a data-driven method that offers an alternative approach to the data association task. In particular, we leveraged the Transformer architecture, introduced in \cite{attention_is_all}, to formulate an assignment decision network, named Transformer-based Assignment Decision Network (TADN), that offers an alternative for common data association techniques that may require an optimization step during inference, such as Linear Data Association Problem (LDAP). Our work is further motivated by the fact that most LDAP solvers such as the Hungarian algorithm are not differentiable, except a few approaches such as the one introduced in (\cite{how-to-train}), thus hindering an end-to-end trainable MOT tracker formulation. Moreover, we discuss the actual contribution of data association in a tracking-by-detection scheme offering several experiments and valuable ablations. Furthermore, in order to demonstrate the effectiveness of our method, we built an online MOT algorithm that integrates the novel data association method. Unlike recent Transformer-based approaches for tracking like MOTR (\cite{motr}) or TrackFormer ({\cite{trackformer}}) that opt for a joint detection and tracking scheme, our approach is based on tracking-by-detection paradigm acting upon a set of predetermined detections.

We conducted experiments on challenging MOT benchmarks such as MOT17 (\cite{MOT16}, \cite{motchallenge-2021}), MOT20 (\cite{mot20}) and UA-DETRAC (\cite{UA-DETRAC}). Our results showed strong, superior even, performance against well known methods varying by their data association approach, even without incorporating any occlusion handling or re-identification components as many of the compared methods do. Our contributions are summarized as follows:
\begin{itemize}
  \item We introduce a novel data association method, named TADN, for visual, online MOT as an alternative to explicitly solving an optimization problem, such as LDAP, during inference. 
  \item We provide a baseline tracking-by-detection implementation along with the source code that integrates TADN coupled with simple submodules for motion and appearance modeling.
  \item We offer an extensive experimental validation in challenging MOT benchmarks that demonstrate the superior performance of TADN against methods employing alternative data association methods.
\end{itemize}

This paper is structured as follows : Section 2 contains a review of the related literature regarding data association methods for multiple object tracking and Transformer architectures. In Section 3 we introduce our novel method for data association, while describing also the implemented tracking algorithm. In Section 4 we submit the results of our approach on two challenging benchmarks along with a detailed ablation study for our method. Last, in Section 5 a series of conclusions and limitations are drawn along with some further future extensions are discussed.

\section{Related Work}
\textbf{Multiple Object Tracking (MOT).} Recent advances of object detection methods (\cite{faster-rcnn}, \cite{detr}) have established tracking-by-detection as a common paradigm for multiple object tracking. Based on the detected objects of interest the goal is to associate them through time in order to form target trajectories. This association process is basically a joint optimization problem where information about newly observed objects is compared to relevant information about active targets in the current frame. Recent works such as DeepMOT (\cite{how-to-train}), BLSTM-MTP (\cite{BLSTM}) and DMAN (\cite{dman}) incorporate an RNN-based model that processes the detected bounding boxes across frames to generate robust object trajectories and leverages the temporal dependencies between consecutive frames to associate the detections and maintain the consistency of object tracks over time. Alternative methods formulate the tracking problem as an energy minimization task on a graph representation in a tracking-by-detection framework (\cite{lifted-multicut}), while other works extend the scope of MOT by treating it as a generalized form of Single Object Tracking (SOT) where target locations are estimated by utilizing an ensemble of SOT models (\cite{chu2017online}, \cite{chu2019online}).

Another popular MOT paradigm is the joint detection and tracking scheme, where both detection and tracking are performed simultaneously in a single stage. A common way to achieve this is to build a tracking-related branch upon an object detector predicting either object tracking offsets or re-identification embeddings for data association (\cite{wang2020towards}). For example, CenterTrack (\cite{centertrack}) utilizes a center point detector to estimate the object centers and combines it with a deep association network to perform online data association, while FairMOT (\cite{fairmot}) utilizes a shared backbone to learn the object detection and appearance embedding task.

In either case, the primary sources of information typically used when determining the matching cost or dissimilarity between detections are the appearance and motion of the object. Motion modeling represents the movement patterns of objects over time to effectively track their trajectories in order to predict their future positions and improve the accuracy of MOT algorithms. Constant velocity assumption (\cite{cva}), Kalman filtering (\cite{kalman1960new}) and probabilistic modeling via gaussian mixture models (\cite{gmm}) are amongst the most standard methods for trajectory prediction, while in recent years deep learning techniques modeling complex motion patterns directly from the data and producing socially acceptable trajectories have been introduced (\cite{socialgan}, \cite{psaltatsir}). On the other hand, appearance modeling plays a significant role in MOT by enabling the discrimination and tracking of objects based on their visual characteristics. Such modeling techniques involve capturing variations in the appearance of objects across frames and handle challenges like illumination changes, occlusions and object appearance variations. Appearance driven tracking methods may rely on image recognition backbones produced by Siamese networks (\cite{leal2016learning}), capitalize on learnable re-identification features (\cite{tracktor}) or combine various cues such as location, topology and appearance (\cite{xu2019spatial}).

Our method follows a tracking-by-detection scheme to better highlight the effect of our novel data association proposition.   We opted for a rather simple configuration regarding appearance and motion modeling to test our method's ability to associate detections to targets without adding any extra computational complexity or unwanted bias that might hinder the interpretation of our output results. Tracking-by-detection methods due to separately conducting detection and tracking tend, in general, to be more computationally expensive. Instead, our approach leverages efficient neural network architectures to perform real-time association in an end-to-end fashion.

\begin{figure*}
  \includegraphics[width=\textwidth]{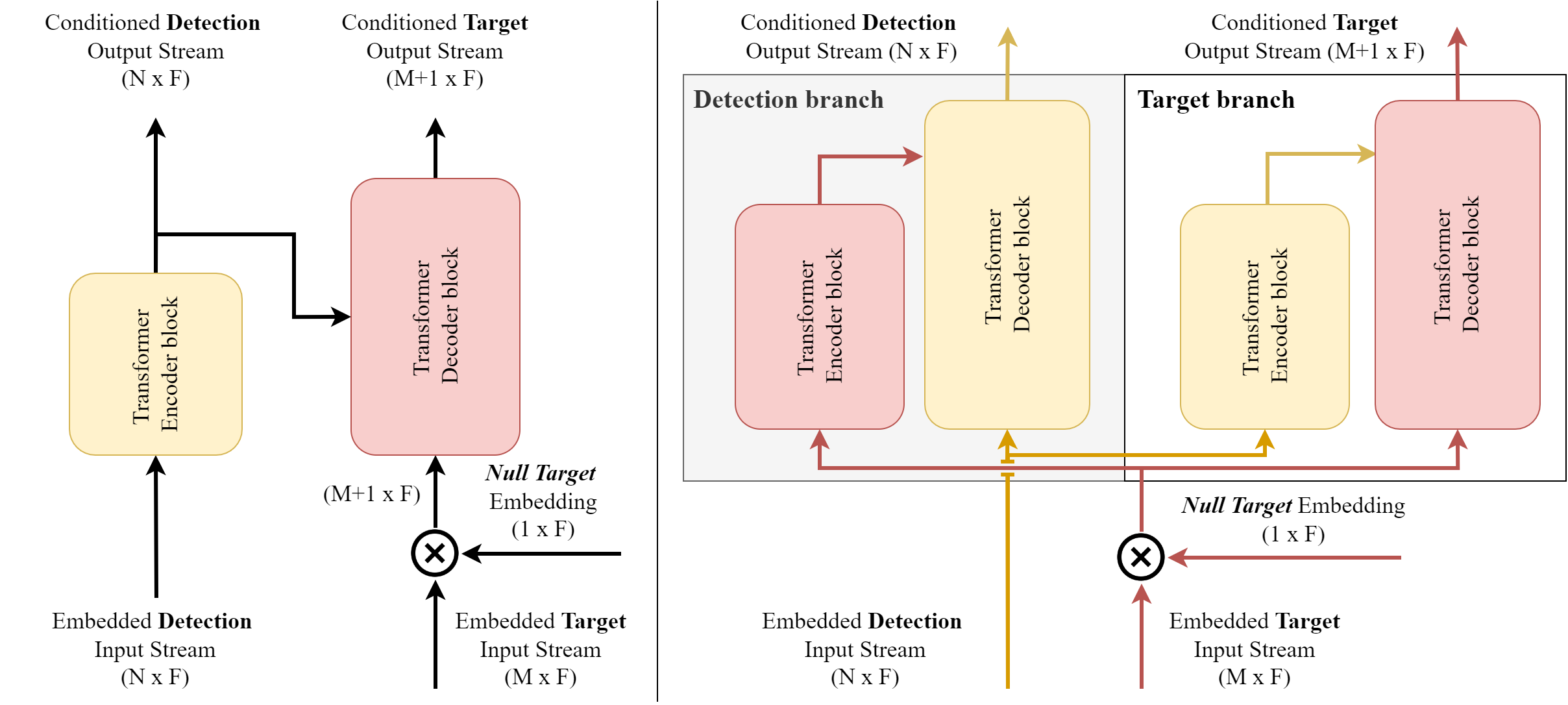}
  \caption{\textbf{Single and dual branch configurations for TADN. Left:} Single branch approach is constituted of a single Transformer model. Target and Detection input streams are fed to the decoder and encoder part respectively. Target output stream is the Transformer's standard output, while Detection output stream the output of the Transformer's encoder. \textbf{Right:} Dual branch version uses two separate Transformer models. Each output stream corresponds to the output of each Transformer model. Detection and Target input stream are fed to the decoder and encoder part respectively for the Detection branch and vice-versa for the Target branch. In both architectures, a $null\; target$ embedding is concatenated to the Target input stream before feeding it to the Transformer.}\label{fig_TADN}
\end{figure*}

\textbf{Data association in MOT.} Data association can be performed offline (batch-mode) by exploiting information both from past and future trajectories (\cite{tang2016multi}). However, the rise of real-time applications have established online methods that exploit only information from past trajectories as the most popular over the years (\cite{chu2019online}, \cite{chu2017online}). Under an online context, the assignment problem can be formulated as a LDAP optimization problem between two consecutive frames and is solved by bipartite matching algorithms such as the Hungarian algorithm (\cite{bewley2016simple-hungarian},  \cite{fang2018recurrent-hungarian}). Building upon that concept, Deep Affinity Network (DAN) introduced by \cite{deep-affinity-network} models both appearance and affinity between objects over time producing a soft assignment matrix and applies the Hungarian algorithm to predict the optimal assignments in a supervised manner using ground truth assignment matrices. Similarly, \cite{famnet} expand on the idea of affinity estimation by directly learning to predict the correct assignments. Moreover, DeepMOT (\cite{how-to-train}) introduces an end-to-end training framework that uses differentiable versions of MOTA and MOTP (\cite{clear-metrics}) evaluation metrics as loss function and proposes the Deep Hungarian Network (DHN) as a linear assignment method between detection hypotheses and active trajectories. 

Alternative data association methods apply under the multiple hypotheses tracking (MHT) scheme, such as \cite{mht-first-paper}, where a trade-off between maintaining and removing hypotheses with high and low confidences respectively is considered. In this case the assignment problem is formulated as a maximum weighted independent set (MWIS) problem. Due to the exponential growth of hypotheses, recent works such as \cite{mht-1} introduce an enhancing detection model to cope with data association ambiguities, while \cite{mht-2} propose a tracklet-level association pruning methods and approximation algorithms for the MWIS problem.

In addition, a significant number of alternative data association approaches rely on network flow-based (\cite{network-flow-1}, \cite{network-flow-3}) or graph matching (\cite{graph-based}) methods where the total sum of pairwise costs is minimized. These assignment costs can either be based on heuristics or even be learnable. \cite{network-flow-2} introduce an end-to-end trainable pipeline for network-flow formulations where a set of arbitrary parameterized cost functions are learned. \cite{lifted-multicut} propose a minimum cost lifted multicut problem as a way to jointly cluster detections hypotheses over time. Also, \cite{heterogeneous-graph} introduce a heterogeneous association fusion (HAF) tracker to deal with detection failures and considers a multiple hypotheses formulation for the assignment problem. Other approaches use reinforcement learning techniques for data association, such as the work from \cite{reinforcement-learning}, where a target trajectory is initialized, terminated or maintained regarding a learnable policy over a Markov decision process.

All in all, most of the approaches discussed above rely on the computation of a cost matrix to explicitly solve a linear data association problem (LDAP). In general, such an approach introduces a computational overhead to a MOT algorithm that varies given the number of detections and targets in a scene. Thus, this process occupies a significant portion of the available computational budget for real time applications. Multiple hypotheses techniques have in principle high computational complexity, while network flow and graph-based data association methods need to enforce global optimization on large graphs constraining their potential for real-time MOT. Our method follows the tracking-by-detection paradigm and is based on an assignment decision network that directly generates assignments during inference instead of relying on solving such optimization problems.

\textbf{Transformer architectures.} Since their first appearance in natural language processing, Transformer networks (\cite{attention_is_all}) have achieved significant success in a plethora of computer vision applications, such as person re-identification (\cite{reID}) and object detection (\cite{detr}). Recent works have successfully investigated the exploitation of these architectures for MOT. TrackFormer (\cite{trackformer}) set up a joint detection and tracking pipeline modifying the object detector DETR (\cite{detr}) to an end-to-end MOT algorithm through propagating ``track" queries over time. In a similar manner, MOTR \cite{motr} formulate MOT as a problem of "set of sequence" predictions and also builds upon DETR (\cite{detr}) extending its object queries to "track queries" to predict object sequences. TransMOT introduced by \cite{transmot} treats target trajectories as a set of sparse weighted graphs and models their interactions through temporal and spatial transformer encoder and decoder layers respectively. TransCenter (\cite{transcenter}) handles detection related problems in tracking, such as missing detections or very dense detections in overlapping anchors, proposing dense detection queries, as well as introduces sparse tracking queries to overcome computational efficiency problems related to the quadratic complexity nature of Transformers. It should be noted that most of these approaches operate directly on the raw image data, thus following a joint detection and tracking paradigm. 

In order to exploit the capacity of these architectures, in this work we introduce a novel data association method based on a Transformer architecture showing strong performance on popular MOT benchmarks. Our intuition is based on the insight that these architectures leverage self-attention and cross-attention mechanisms to model the relationships between the encoder and decoder inputs, that in our case are mapped to detection boxes and existing trajectories in the scene respectively. Our method differs significantly from other Transformer-based MOT algorithms by designing a Transformer-based association method operating on preexisting detections provided by any arbitrary object detector. In contrast, other methods, such as TrackFormer or MOTR, build and adapt upon the well known Transformer-based object detector DETR resulting in a joint detection and tracking solution, while ours is purely a tracking-by-detection approach.

\begin{figure*}
  \includegraphics[width=\textwidth]{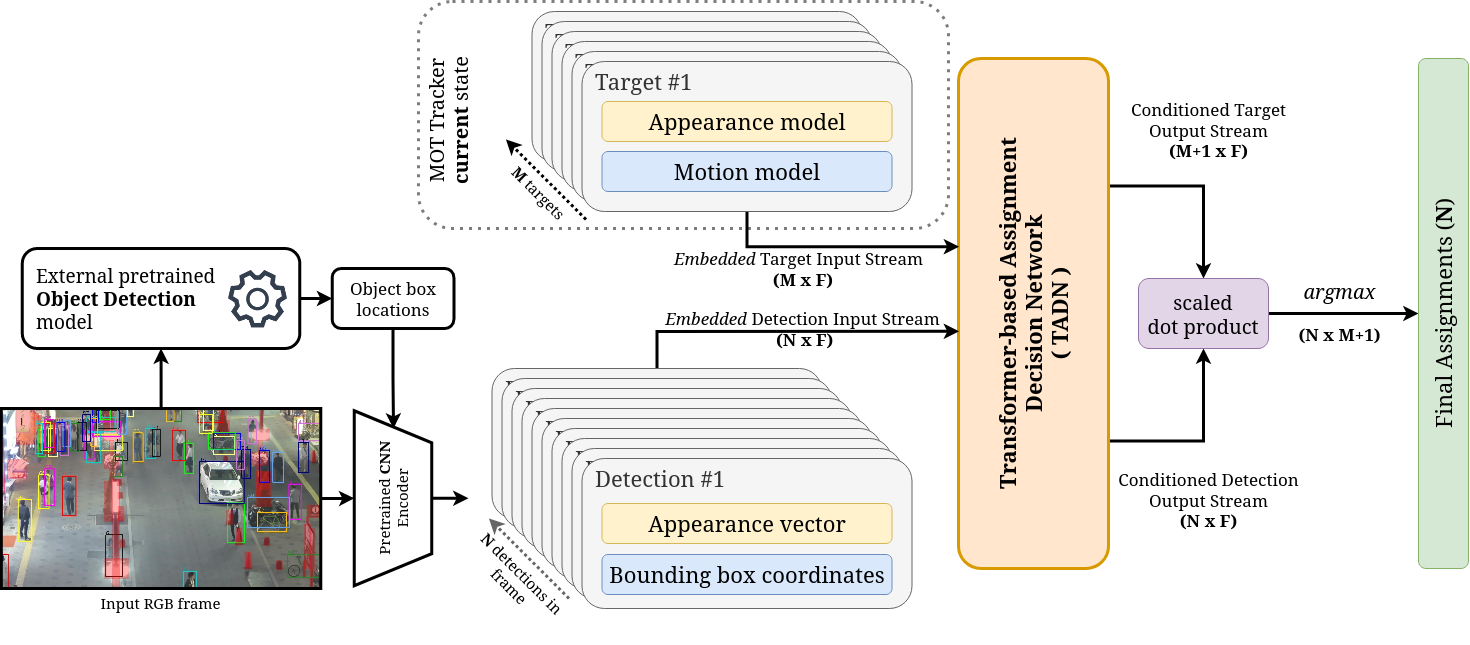}
  \caption{\textbf{Overview of our MOT pipeline}. Two input streams containing positional and appearance information are generated for $N$ detections and $M$ currently active targets respectively. These are fed to TADN to compute an $(N \times M+1)$ similarity. Final assignments are directly computed via a row-wise \emph{argmax} operation. The final assignments include the $null\; target$ case.}
  \label{Fig1}
\end{figure*}

\section{Methodology}\label{sec3}

We  designed and implemented Transformer-based Assignment Detection Network (TADN) as an alternative to data association for MOT. In short, TADN transforms information related to detections and known targets in each frame to directly compute optimal assignments for each detection. To accomplish this an external object detector is needed to compute detection locations for each frame, while in contrast, other Transformer-based approaches such as TrackFormer (\cite{trackformer}) act upon the original image frame to simultaneously generate detections and assignments. To evaluate TADN's real-world performance, a simple tracking-by-detection MOT algorithm was implemented to support it. Our tracker, including TADN, was end-to-end trained via the proposed novel training strategy.

\subsection{Transformer-based Assignment Detection Network}

TADN operates on every time-step to condition two separate input data streams using a series of self- and cross-attention mechanisms. Let $D_{in} \in \mathbb{R} ^{N \times F_D}$ be the set of $N$ detections in each frame while $T_{in} \in \mathbb{R} ^{M \times F_T}$ be the set of $M$ previously tracked targets.

In either stream, information from multiple cues is incorporated into a set of vectors via linear embedding layers $L_D, L_T$  of dimension $d_{model}$ producing $D'_{in} \in \mathbb{R} ^{N \times d_{model}}$ and $T'_{in} \in \mathbb{R} ^{M \times d_{model}}$ as  shown in Equation \ref{eq_lin_proj}.

\begin{equation}
\label{eq_lin_proj}
\begin{aligned}
D'_{in}  = L_D (D_{in}) \in \mathbb{R} ^{N \times d_{model}}\\
T'_{in}  = L_T (T_{in}) \in \mathbb{R} ^{M \times d_{model}}
\end{aligned}
\end{equation}

The final output of the TADN module is two sets of vectors. Let $D_{out} \in \mathbb{R}^{N\times  d_{model}}$ and $T_{out} \in \mathbb{R}^{M\times  d_{model}}$ be the set of output transformed feature vectors related to detections and actively tracked targets respectively:

\begin{equation}
\label{eq_tadn_ovr}
D_{out}, T_{out} = TADN(D'_{in}, T'_{in};\theta_{TADN})
\end{equation}
where $\theta_{TADN}$ the set of TADN's architecture parameters.

It should be noted that $ T_{out}$ has one more vector compared to the corresponding input set $T_{in}$ to compensate for the case of not assigning a detection to a specific target. This extra target is identified as the $null\;target$ and a corresponding learnable embedding vector $T_{null} \in \mathbb{R} ^{d_{model}}$ is concatenated to $T'_{in}$. In that way, Equation \ref{eq_lin_proj} is adjusted as:

\begin{equation}
\label{eq_lin_proj_adj}
\begin{gathered}
D'_{in}  = L_D (D_{in}) \in \mathbb{R} ^{N \times d_{model}}\\
T'_{in}  = L_T (T_{in}) \oplus T_{null} \in \mathbb{R} ^{M+1 \times d_{model}}
\end{gathered}
\end{equation}
where $\oplus$ denotes the concatenation operator.

To derive the final assignments, first the $Assignment\;Score\;Matrix$ $(ASM)$ is computed as the scaled dot product between the two output sets:

\begin{equation}
\label{eq_asm}
ASM_{N\times M+1} = \frac{D_{out}T_{out}^T}{\sqrt{d_{model}}}\
\end{equation}

Final assignments $A_{final}$ are computed as the $argmax$ of each row of the $ASM$ as shown in Equation \ref{eq_final_ass}. Assigning a detection to the last column of $ASM$ corresponds to an assignment to the $null\;target$ i.e. a non-assignment.

\begin{equation}
\label{eq_final_ass}
\begin{gathered}
A_{final} = \mathbf{argmax}(ASM, dim=1) \in \mathbb{I}^{N}\\
where \; \mathbb{I} = \{1, ... , M+1\} \subset \mathbb{N}
\end{gathered}
\end{equation}

In detail, TADN exploits the set-to-set nature of Transformers. In fact, two different architecture configurations have been designed for TADN; a \emph{single} $TADN_{single}$ and a \emph{dual} $TADN_{dual}$ branch version as shown in Figure \ref{fig_TADN}. 

Regarding $TADN_{single}$ (Figure \ref{fig_TADN}, Left), $D'_{in}$ and $T'_{in}$ are fed directly into the encoder and decoder part respectively of a typical Transformer model introduced by \cite{attention_is_all}. $T_{out}$ and $D_{out}$ are mapped to the outputs of the decoder and encoder parts  of the Transformer respectively.

\begin{equation}
\label{eq_single_tadn}
\begin{gathered}
D_{out} = \mathbf{Enc}(D'_{in}; \theta_{enc})\\
T_{out} = \mathbf{Dec}(T'_{in}, D_{out}; \theta_{dec})\\
\end{gathered}
\end{equation}
where $\mathbf{Enc}$ and $\mathbf{Dec}$ are the Encoder and Decoder parts of a typical Transformer architecture with $\theta_{enc}$ and $\theta_{dec}$ parameters respectively.

Here, $D_{out}$ is solely dependent on the $D'_{in}$ since the encoder part of the transformer is comprised only from self-attention layers, while $T_{out}$ is jointly dependent on both $D'_{in}$ and $T'_{in}$ since apart from the self-attention layers, the decoder part has one cross-attention layer for each decoder layer. Our main intuition is that $T_{out}$ contains information about the active targets transformed by taking also into consideration the detections found in the next frame of the sequence. On the other hand, $D_{out}$ contains information directly derived from $D'_{in}$ transformed in order to be useful in the cross-attention layer of the decoder.

For $TADN_{dual}$, two separate Transformer models are used (Figure \ref{fig_TADN}, Right). Each Transformer outputs either $D_{out}$ or $T_{out}$. In the \emph{detection branch} $T'_{in}$ is fed in the encoder while $D'_{in}$ in the decoder of the Transformer to derive $D_{out}$ and vice-versa for the \emph{target branch} to derive $T_{out}$:

\begin{equation}
\label{eq_dual_tadn}
\begin{gathered}
D_{out} = \mathbf{Transformer_D}(T'_{in}, D'_{in}; \theta_{D})\\
T_{out} = \mathbf{Transformer_T}(D'_{in}, T'_{in}; \theta_{T})\\
\end{gathered}
\end{equation}
where $\mathbf{Transformer_D}$ and $\mathbf{Transformer_T}$ are the Transformer models for  \emph{detection} and \emph{target} branch with parameters $\theta_{D}$ and $\theta_{T}$ respectively.

In that case, all known targets attend to all detections in either branch. $TADN_{dual}$ requires essentially double learnable parameters and computations than $TADN_{single}$. By intuition $TADN_{dual}$ has higher capacity than $TADN_{single}$ since both $T_{out}$ and $D_{out}$ are dependent on $T'_{in}$ and $D'_{in}$ as cross-attention layers are found in either branch, whereas in the \emph{single branch} version only $T_{out}$ is conditioned on both inputs.

\subsection{Tracking-by-detection with TADN}

To demonstrate our method’s performance we opted for a tracking-by-detection scheme featuring only an appearance model and a motion model to generate representations for detections and targets (Figure \ref{Fig1}). As a motion model we opted for the standard Kalman filter (\cite{kalman1960new}) paired with Enhanced Correlation Coefficient (ECC) based motion compensation introduced by \cite{ecc}. As an appearance model we chose to retain raw CNN features for the last assignment on a patch-level from a ResNet50 (\cite{resnet50}) architecture pretrained on a re-identification downstream task, as described in \cite{tracktor}. However it should be noted that ResNet50 is used exclusively as a feature extractor in our algorithm and not as part of a re-identification module where target re-birth can occur.  No further specialized modules such as re-identification (Re-ID) or occlusion handling have been included in our tracking algorithm. We opted for such a basic configuration to remove any bias/gains induced from using more specialized components. Doing so, we validated the performance impact of TADN in an unbiased way. 

As described in Algorithm \ref{alg_assign}, our MOT pipeline leverages information generated from each incoming frame along with information regarding actively tracked targets already known from previous steps. More specifically, using a pretrained object detector box locations are derived for the objects in the scene. For each detected object, its corresponding image patch is extracted and CNN features are computed. These appearance and positional features are concatenated into $D_{in}$. Similarly, for each active target its positional and appearance features are concatenated also into $T_{in}$. Subsequently, $D_{in}$ and $T_{in}$ are fed to TADN which outputs $ASM$. Final assignments are derived using Equation \ref{eq_final_ass}.

\begin{algorithm}
 \caption{Compute assignments}\label{alg_assign}
 \begin{algorithmic}[1]
 \renewcommand{\algorithmicrequire}{\textbf{Input:}}
 \renewcommand{\algorithmicensure}{\textbf{Output:}}
 \REQUIRE Tracker state: motion models $\{MM_i\}_{i=1}^M$ 
 \newline Tracker state: appearance models $\{AM_i\}_{i=1}^M$,
 \newline Image frame $\mathbf{I} \in \mathbb{R}^{H\times W\times 3}$,
 \newline Detection locations $\{BB_i\}_{i=1}^N,\;BB_i \in \mathbb{R}^4$,
 \newline CNN feature extractor $\mathbf{F}:\mathbb{R}^{H\times W\times 3} \rightarrow \mathbb{R}^{d_{F}}$
 \ENSURE  Final assignments for frame $\mathbf{I}$ 
 \newline  \textit{Compute $D_{in}$} :
  \STATE $D_{in} \gets \emptyset$
  \FOR {each detection $i = 1, ..., N$}
  \STATE Get location $BB_i = (x_i, y_i, w_i, h_i)$
  \STATE Crop patch $P_i \gets \mathbf{I}(BB_i)$
  \STATE Compute appearance features \newline $A^D_i \gets \mathbf{F}(P_i)$
  \STATE Update $D_{in} \gets D_{in} \cup \{ L_{D}(A^D_i \oplus  BB_i) \}$
  \ENDFOR
 \newline  \textit{Compute $T_{in}$} :
 \STATE $T_{in} \gets \emptyset$
 \FOR {each target $j = 1, ..., M$}
 \STATE Predict location $BB_j \gets MM_j.predict()$
 \STATE Retrieve appearance features\newline$A^T_j \gets AM_j.step()$
 \STATE Update $T_{in} \gets T_{in} \cup \{ L_{T}(A^T_j \oplus BB_j) \}$
 \ENDFOR
 \newline  \textit{Compute assignments} :
 \STATE $D_{out}, T_{out} \gets \mathbf{TADN}(T_{in}, D_{in}; \theta_{TADN})$
 \STATE Compute $\mathbf{ASM}$ as in Equation \ref{eq_asm}
 \STATE $A_{final} \gets \mathbf{argmax}(\mathbf{ASM}, dim=1)$
 \newline
 \RETURN $A_{final}$ 
 \end{algorithmic} 
 \end{algorithm}

To enable complete tracking functionality, we opted for a set of heuristic rules regarding post-association steps such as target update, spawn and termination (Algorithm \ref{alg_update}). Initially, given that TADN outputs one assigned target (including the $null\;target$) per detection, we filter-out any duplicate targets in the assignments (excluding $null\;target$) based on highest prediction confidence. Any filtered-out assignments are discarded. The remaining are used to update the motion and appearance models of the assigned targets. Next, the target termination step is performed; any targets that haven't been assigned to any detection for $T_h$ past consecutive frames are set to an inactive state and their trajectory is considered final. However, having a constant threshold $T_h$ is problematic since we need to terminate early those trajectories with very few ``hits", i.e. total detections assigned to target since its creation. On the other hand, we can rely more on the motion model trajectory predictions for targets that have a long history, thus we opted for a variable $T_h$ as a function of the total number of ``hits" per target. We used a sigmoid function that smoothly transitions from $T_h^{min}$ to $T_h^{max}$ as number of ``hits" rises from 0 to $H_{max}$:

\begin{equation}
\label{eq_sigmoid}
\begin{gathered}
t_\delta = T_h^{max} - T_h^{min}\\
T_{h_i} = t_\delta \cdotp \sigma(15\; (\frac{h_i}{H_{max} - 0.5}) + T_h^{min})
\end{gathered}
\end{equation}
where $\sigma$ is the sigmoid function, $h_i$ the total number of hits for target $i$.

Finally, new targets are spawn given all $null\;target$-assigned detections. For each such detection a new tracklet is created by initializing its motion and appearance models using the corresponding box locations and CNN features.

\begin{equation}
\label{eq_sigmoid}
\begin{gathered}
t_\delta = T_h^{max} - T_h^{min}\\
T_{h_i} = t_\delta \cdotp \sigma(15\; (\frac{h_i}{H_{max} - 0.5}) + T_h^{min})
\end{gathered}
\end{equation}
where $\sigma$ is the sigmoid function, $h_i$ the total number of hits for target $i$.

Finally, new targets are spawn given all $null\;target$-assigned detections. For each such detection a new tracklet is created by initializing its motion and appearance models using the corresponding box locations and CNN features.

\begin{algorithm}
 \caption{Target management}\label{alg_update}
 \begin{algorithmic}[1]
 \renewcommand{\algorithmicrequire}{\textbf{Input:}}
 \renewcommand{\algorithmicensure}{\textbf{Output:}}
 \REQUIRE Assignments $A_{final} \in \mathbb{R}^N$
 \newline  Assignment scores $\{S_i\}_{i=1}^N, S_i \in \mathbb{R}$
 \newline Tracker state: motion and appearance models $\{MM_i, AM_i\}_{i=1}^M$
 \newline Detection locations $\{BB_i\}_{i=1}^N,\;BB_i \in \mathbb{R}^4$,
 \newline Detection appearance $\{A_i^D\}_{i=1}^N, A_i^D \in \mathbb{R}^{F_D}$
 \ENSURE  Updated tracker state
 \newline  \textit{Filter-out duplicate assignments} :
  \STATE $A_{valid} \gets \emptyset$
  \FOR {each detection $D_i \; i = 1, ..., N$}
  \STATE Get assigned target $T^{assign}_i \gets A_{final}[i]$
  \IF {($T^{assign}_i$ is $null\;target$)}
  \STATE $A_{valid} \gets A_{valid} \cup \{(i, T_{null}, -1)\}$
  \ELSIF{($T^{assign}_i$ not in $A_{valid}$)}
  \STATE $A_{valid} \gets A_{valid} \cup \{(i, T^{assign}_i, S_i)\}$
  \ELSE {}
  \STATE $D_e, _, S_e \gets A_{valid}[$target is $T^{assign}_i]$
    \IF {($S_i > S_e$)}
    \STATE $A_{valid} \gets A_{valid}  - \{(D_e, T_e, S_e)\} $
    \STATE $A_{valid} \gets A_{valid} \cup \{(i, T^{assign}_i], S_i)\} $
    \ENDIF
  \ENDIF
  \ENDFOR
  \newline  \textit{Update, terminate and spawn targets} :
  \FOR {each target $T_i,\; i = 1, ..., M$}
  \IF {($T_i$ in $A_{valid}$)}
  \STATE $D_i, _, S_i \gets A_{valid}[$target is $T_i]$
  \STATE Update target $(MM_i, AM_i)$\newline using $BB_{D_i}, A_{D_i}^D$
  \ELSE
  \STATE Propagate target state $(MM_i, AM_i)$
  \ENDIF
  \STATE $T_{h_i} \gets$ \# of $hits$ for $T_i$
  \IF {($T_i$ \# steps unassigned $> T_{h_i}$)}
  \STATE $T_i \gets$ inactive
  \ENDIF
  \ENDFOR
  \STATE $M^* \gets M$
  \FOR {each detection $D_i,\;i = 1, ..., N$}
  \IF {($A_{valid}[D_i]$ contains $T_{null}$)}
  \STATE Initialize a new target $k=M^*+1$\newline $(MM_k, AM_k)$ using $BB_i, A_i^D$
  \STATE $M^* \gets M^* + 1$
  \ENDIF
  \ENDFOR 
  \newline
 \RETURN Updated tracker state \newline $\{\mathbf{MM_i, AM_i}\}_{i=1}^{M^*}$ 
 \end{algorithmic} 
 \end{algorithm}

\subsection{Training strategy}

To train our MOT model, we designed an end-to-end training strategy. We formulated a classification problem which is optimized for every frame and considers each detection separately. The target categories are the set of active targets plus an extra category for the $null\;target$. Given that the outcome of TADN, i.e. the $Assignment\;Score\;Matrix\;(ASM)$ is a matrix of shape ($N\times M+1$), it is straightforward to serve as the ``logits" output for the classification problem. 

To define a categorical cross-entropy loss function over these outcomes a target one-hot encoded (row-wise) matrix of similar shape was formulated, named $Label\;Assignment\;Matrix$ $(LAM)$ (Algorithm \ref{alg_lam}). To compute $LAM$, first, we associated in the previous frame the set of actively tracked targets $\{T_i^{MOT}\}$ for $i=1,...,M$ to the set of ground truth targets $\{T_i^{GT}\}$ for $i=1,...,M', M\neq M'$ using the Hungarian algorithm and a suitable pairwise distance metric ($pdist$), such as the IoU metric or other, as discussed in the following section:  

\begin{equation}
\label{eq_hungarian}
\begin{gathered}
CM = pdist(\{T_i^{MOT}\}_{i=1}^M, \{T_j^{GT}\}_{j=1}^{M'})\\
\{(T_{i(k)}^{MOT}, T_{j(k)}^{GT})\}_{k=1}^K = H(CM; T_{assign})
\end{gathered}
\end{equation}
where $CM  \in \mathbb{R}^{M \times M'}$ is the pairwise distance matrix, $H$ the Hungarian algorithm.

Next, we enforced a minimum threshold $T_{assign}$ to filter-out the assigned pairs resulting in $K$ pairs. After this step a target $T_i^{MOT}$ can either correspond to a single ground truth target $T_j^{GT}$ (\emph{``on-track" state}) or to none (\emph{``off-track" state}). For \emph{``on-track"} targets we used ground truth box locations since there is a known association, whereas for \emph{``off-track"} we predicted new box locations using the motion model. For the current frame, a ($N \times M$) pairwise distance matrix is computed between these box locations and the detections using the same metric. Using the $argmax$ function we transform each row into a one-hot vector if its max value is over a threshold $T_{det2gt}$, else we zero-out the whole row. An extra column with zeros is appended to compensate for the non-assignment case resulting in a ($N x M+1$) matrix. Last column elements are set to 1 if all other columns are set to 0 for each row. Finally, each row of the resulting $LAM$ matrix is a one-hot vector denoting an association of each detection to a single target (including the $null\;target$ case) and serves as the ``label" for the classification problem.

\begin{algorithm}[t]
 \caption{LAM computation}\label{alg_lam}
 \begin{algorithmic}[1]
 \renewcommand{\algorithmicrequire}{\textbf{Input:}}
 \renewcommand{\algorithmicensure}{\textbf{Output:}}
 \REQUIRE Active targets $\{T_i^{MOT}\}$, $i=1, ..., M$
 \newline Ground-truth targets $\{T_i^{GT}\}$, $i=1, ..., M'$
 \newline Detections $\{D_i\},\;i=1,...,N$
 \newline Target Assignment threshold $T_{assign}$
 \newline Detections assignment threshold $T_{det2gt}$
 \ENSURE  $LAM \in \mathbb{R}^{N\times M+1}$
 \newline  \textit{Compute K pairs of active to ground-truth targets} :
  \STATE  $\{(T_{i(k)}^{MOT}, T_{j(k)}^{GT})\}_{k=1}^K \gets$ Equation \ref{eq_hungarian}
  \newline  \textit{Classify targets} :
  \STATE  \emph{on-track} $\gets \emptyset$
  \STATE  \emph{off-track} $\gets \emptyset$
  \FOR {each active target $T_i^{MOT} i = 1, ..., M$}
  \IF {($T_i^{MOT}$ in $\{(T_{i(k)}^{MOT}\}_{k=1}^K$)}
  \STATE \emph{on-track} $\gets$ \emph{on-track} $\cup \{T_i^{MOT}\}$
  \ELSE {}
  \STATE \emph{off-track} $\gets$ \emph{off-track} $\cup \{T_i^{MOT}\}$
  \ENDIF
  \ENDFOR
  \newline  \textit{Compute bbox locations} :
  \STATE Generate ${BB_i}_{i=1}^M$ by predicting bbox locations for targets in \emph{off-track} and using GT bbox locations for targets in \emph{on-track}.
  \newline  \textit{Associate detections to targets} :
  \STATE $LAM \gets pdist({D_i}_{i=1}^N,{BB_j}_{j=1}^M)$
  \STATE $LAM[LAM < T_{det2gt}] \to 0$
  \STATE $LAM[LAM \neq max(LAM, dim=1)] \to 0$
  \STATE $C_1 \gets LAM = max(LAM, dim=1)$
  \STATE $C_2 \gets LAM \neq 0$
  \STATE $LAM[C_1\cap C_2] \to 1$
  \newline  \textit{Append a column for} $null\;target$ :
  \STATE $NT \gets \mathbb{O}^{N \times 1}$
  \STATE $NT[max(LAM, dim=1) = 0] \to 1$
  \STATE $LAM \gets LAM \oplus NT$
  \newline 
 \RETURN $LAM$
 \end{algorithmic} 
 \end{algorithm}

Having computed $LAM$ and estimated $ASM$ using TADN, we define a categorical cross-entropy loss between them to train our model on. To compensate for the variable number of active targets during training that directly impacts the total categories in the classification problem, we normalized the value of the loss function by dividing with the total number of active targets at each frame. Ultimately, loss function is mean-reduced for all detections:

\begin{equation}
\label{eq_loss}
{\mathcal{L}} = \frac{1}{N\cdotp (M+1)} \cdotp \displaystyle\sum_{i=1}^{N}\displaystyle\sum_{j=1}^{M} LAM_{i,j} \cdotp log (ASM_{i,j}) 
\end{equation}

After all optimization steps have taken place, we update all targets as described in the previous subsection. However, since this update process is based on the model's predictions, it is expected that in early training stages a lot of erroneous assignments occur due to our model's random initialization. This could lead to an evolving set of targets that wouldn't be representative of a real world tracking problem and could potentially hinder or slow down training. To overcome this, we opted to use \emph{LAM} to generate assignments instead of the predicted ones. This only affects the latter part of the training step and not the actual optimization procedure. As training epochs progress we used a ``choice" probability $p_{choice}$ in order to choose between \emph{LAM}-generated and TADN-predicted assignment for each detection individually. We gradually altered $p_{choice}$ in favor of our model's prediction using a sigmoid-based transition function relative to elapsed number of epochs. After a certain epoch $E_{max}$, we used our model's prediction exclusively to step our tracking algorithm:

\begin{equation}
\label{eq_pchoice}
p_{choice}(e) = \sigma(-\frac{c}{2} + c \; \frac{e-E_{min}}{E_{max}-E_{min}}) 
\end{equation}
where $e$ the current epoch, $\sigma$ the sigmoid function, $E_{min}, E_{max}$ the starting and ending epoch for the transition, while $c$ a scaling factor typically set to 12.

\section{Experimental Results}\label{sec4}

We conducted a series of experiments on three well-known datasets, namely MOT17 (\cite{MOT16}), MOT20 (\cite{mot20}) and UA-DETRAC (\cite{UA-DETRAC}). TADN performed well against all directly comparable methods in many significant metrics, especially delivering superior MOTA performance, thus proving to be a viable alternative to typical computationally expensive data association techniques. In this section, first we present the benchmarks we evaluated our method on followed by our implementation details. Next, we present our results along with a qualitative analysis including particular cases of interest regarding the strengths and weaknesses of our method. Last, an extensive ablation study follows investigating and comparing the performance of our method for a curated selection of various configurations.

\subsection{Datasets and metrics}

\textbf{MOTChallenge}. MOT17 and MOT20 datasets are part of MOTChallenge (\cite{MOT16}) which is a standard pedestrian tracking benchmark suite. MOT17 consists of 7 train and 7 test video sequences depicting stationary and moving pedestrians on a variety of scene configurations varying in camera movement, point of views, occlusions, lighting conditions and video resolutions. Under a static camera configuration, MOT20 consists of 4 train and 4 test heavily crowded and challenging scenes. All these variations pose a significant challenge for multiple object tracking thus increasing the popularity of MOTChallenge for measuring the performance for such algorithms. MOT17 dataset provides a public detection track provided by three detection algorithms, i.e. Faster-RCNN (\cite{faster-rcnn}), DPM (\cite{DPM-detector}) and SDP (\cite{SDP-detector}), while MOT20 provides only Faster-RCNN detections. To this end, the performance of MOT algorithms between those exploiting either public or private detections are distinguished on separate tracks.

For evaluation purposes this benchmark utilizes CLEAR metrics (\cite{clear-metrics}). MOTA metric combines three error ratios, i.e. false positives, identity switches and missed targets, to intuitively measure accuracy and is considered among the most important metrics for evaluating the performance of a MOT algorithm. On the other hand, MOTP highlights the ability of an algorithm to estimate precise object locations, while IDF1 shows a measure of how many detections are correctly identified. Also, CLEAR metrics provide a thorough insight on how many times a trajectory is fragmented (Frag), how many trajectories are covered by a track hypothesis for their life span (MT, ML) and the total number of missed targets (FN) and identity switches (IDSW).

\textbf{UA-DETRAC}. The UA-DETRAC challenge (\cite{UA-DETRAC}), \cite{UA-DETRAC-2}, \cite{UA-DETRAC-3}) is a vehicle tracking benchmark that includes a mixture of different traffic, weather and illumination conditions from real-world traffic scenes. This benchmark consists of 60 train sequences and 40 test sequences with rich annotations and contains fixed-size sequences (960 x 540) that are recorded at 25 fps with a static camera configuration. Due to its large size and variety of traffic conditions, UA-DETRAC is in fact a very challenging dataset suitable for demonstrating the performance of state-of-the-art MOT algorithms. The benchmark provides a set of public detections from CompACT (\cite{CompACT-detector}), DPM (\cite{DPM-detector}), R-CNN (\cite{rcnn-detector}) and ACF (\cite{acf-detector}), while detections from the benchmark's detection track are also publicly available such as EB (\cite{EB-detector}).

In order to consider both detection and tracking during evaluation, this challenge expands upon CLEAR metrics using the Precision-Recall (PR) curve that is generated by gradually altering the threshold referring to the confidence of the prediction by the object detector. The average of each metric from these series of results for different threshold values is computed, thus introducing the official evaluation metrics of UA-DETRAC, namely PR-MOTA, PR-MOTP, PR-MT, PR-ML, PR-IDSW, PR-FN and PR-FP. It is clear that these metrics are not the most suitable to evaluate our method due to their strong dependency on the selected detector. To mitigate this and provide a detector agnostic performance measure, the maximum MOTA attained was also reported.

\subsection{Implementation details}

We trained our model on the full MOT17 train set for 300 epochs and on the MOT20 train set for 50 epochs using a learning rate of $1e^{-4}$ with a step learning rate scheduler that decreased the learning rate value by a factor of $0.1$ after 180 and 50 epochs respectively. For UA-DETRAC we trained our model on the full train set for 24 epochs and decreased the learning rate after 15 epochs. On either case we set $T_h^{min} = 3$,  $T_h^{max} = 30$, $H_{max} = 100$ and we used Adam optimizer (\cite{adam}). For MOT17 and MOT20 we used and $E_{min} = 20$, $E_{max} = 160$ and for UA-DETRAC $E_{min} = 2$, $E_{max} = 10$. For all datasets we chose to filter-out any detections that have confidence less than 0.3, while also we normalized the bounding box coordinates relative relative to the corresponding image size. We opted for a batch size of 64. It is important to note that our training strategy processes each sample sequentially due to tracking information that need to be updated and propagated across each consecutive sample. Thus, to update our network's parameters during training we opted for gradient accumulation techniques to form a mini-batch.

Regarding the architectural hyper-parameters, we opted for a dual branch TADN module with 2 encoder and decoder layers with 2 heads for each attention layer exploiting both positional and appearance features. We set $d_{model} = 128$, where 64 correspond to positional $L_{pos}$ and 64 to appearance features  $L_{app}$. We used a MOT17 trained reid ResNet50 (\cite{resnet50}) as a feature extractor and a CMC-enabled Kalman motion model for all datasets. For all sequences, camera motion compensation (CMC) features were computed using the ECC algorithm introduced by \cite{ecc} to compensate for sudden camera motion changes present in some scenes. 

We used the publicly available detections from the SDP detector (\cite{SDP-detector}) to train our models for MOT17 and detections produced from the EB (\cite{EB-detector}) detector for UA-DETRAC. For MOT17 and MOT20 private detection tracks we used community pretrained YOLOX (\cite{yolox}) object detectors as provided by the official implementation of ByteTrack (\cite{bytetrack}). For the former, YOLOX detector is pretrained on CrowdHuman (\cite{crowdhuman}), MOT17, CityPersons (\cite{zhang2017citypersons}) and ETHZ (\cite{li2017scale}), while for the latter on CrowdHuman and MOT20. It should be noted however, that TADN model itself, in any case, was trained exclusively on the benchmark provided training data without using any external tracking dataset. Our method was trained and evaluated on a machine with  Ryzen 9 3900X 12 core @3.7GHz CPU, a Nvidia 2080Ti GPU and 64GB RAM. The proposed approach was implemented on Pytorch.

As an assignment metric during training time, we opted not to use the Intersection over Union (IoU) metric that is commonly used in such scenarios since it has a constant zero value in cases where two bounding boxes do not overlap. Also, its value alternates in a non-linear fashion for changes in the relative positions between bounding boxes. In the early training stages, it is expected of the predicted bounding boxes to have a significant offset from the ground truth boxes, thus the behavior of the IoU metric could produce false or non-assignments for our model to train on yielding worse performance and slowing down the training process. To mitigate this we introduced a custom assignment metric, namely \emph{upper-left bottom-right} $L_1$ ($ulbr_1$):

\begin{equation}
\label{eq_ulbr1}
\mathbf{ulbr_1}(bb_i, bb_j) = - \bigg\lvert \frac{\lVert ul_i - ul_j \rVert _1 + \lVert br_i - br_j \rVert _1}{\lVert ul_i - br_j \rVert _1 - \lVert br_i - ul_j \rVert _1} \bigg\rvert
\bigskip
\end{equation}
where $bb_k = [ul_k, br_k] = [x_{min}^k, y_{min}^k, x_{max}^k, y_{max}^k]$. 

As discussed in following sections, $ulbr_1$ due to its behaviour to smoothly transition its value even for non-overlapping boxes proves to be a more suitable metric compared to $IoU$ for determining detections to targets assignments during training. For our experiments, we opted for $T_{tgt2det} = -0.13$ using $ulbr_1$ as the assignment metric of choice.

\subsection{Quantitative evaluation on Benchmarks}

\begin{table*}
\caption{MOT17 results on public detections}\label{table-resultsMOT17}
\centering
\begin{minipage}{\textwidth}
\resizebox{\textwidth}{!}{
\begin{tabular*}{\textheight}{@{\extracolsep{\fill}}*3l*9c@{\extracolsep{\fill}}}
    \textbf{Tracker} & \textbf{Method} & \textbf{DA} & \textbf{MOTA$\uparrow$} & \textbf{IDF1$\uparrow$} & \textbf{MT$\uparrow$} & \textbf{ML$\downarrow$} & \textbf{FP$\downarrow$} & \textbf{FN$\downarrow$} & \textbf{IDSW$\downarrow$} & \textbf{Frag$\downarrow$} & \textbf{Hz\footnotemark[1]$\uparrow$} \\
    \hline
    TL-MHT (\cite{mht-2}) & Offline & MHT & 50.6 & 56.5 & 17.6 & 43.4 & 22213 & 255030 & 1407 & \textbf{2079} & - \\
    MHT-DAM (\cite{multiple-hypothesis}) & Offline & MHT & 50.7 & 47.2 & 20.8 & 36.9 & 22875 & 252889 & 2314 & 2865 & 0.9 \\
    IQHAT (\cite{iqhat}) & Offline & MAIQP & 58.4 & 61.8 & 24.1 & 35.2 & 15013 & 218274 & \textbf{1261} & - & \textbf{8.1} \\
    SP-CON (\cite{sp-con}) & Offline & - & \textbf{61.5} & \textbf{63.3} & \textbf{26.4} & \textbf{32.0} & \textbf{14056} & \textbf{200655} & 2478 & - & 7.7 \\ 
    DASOT (\cite{dasot}) & Online & LA & 48.0 & 51.3 & 19.9 & 34.9 & 38830 & 250533 & 3909 & -  & 9.1 \\
    DeepMOT (\cite{how-to-train}) & Online & e2e-LA & 48.1 & 43.0 & 17.6 & 38.6 & 26490 & 262578 & 3696 & 5353 & 4.9\\
    DMAN (\cite{dman}) & Online & LA & 48.2 & \textbf{55.7} & 19.3 & 38.3 & 26218 & 263608 & 2194 & 5378 & 0.3 \\
    CoCT (\cite{sheng2020near}) & Online & Blockchain & 50.1 & 55.6 & \textbf{23.9} & 35.4 & 49887 & 229391 & \textbf{2075} & - & \textbf{57.8} \\
    OneShotDA (\cite{oneshotda}) & Online & OSL-MN & 51.4 & 54.0 & 21.2 & 37.3 & 29051 & 243202 & 2118 & \textbf{3072} & 3.4 \\
    BLSTM-MTP-O (\cite{BLSTM}) & Online & GDA & 51.5 & 54.9 & 20.4 & 35.5 & 29616 & 241619 & 2566 & 7748 & 20.1 \\
    FAMNet (\cite{famnet}) & Online & e2e-MDAP & 52.0 & 48.7 & 19.1 & 33.4 & 14138 & 253616 & 3072 & 5318 & $<$0.1 \\
    DAN (\cite{deep-affinity-network}) & Online & e2e-LA & 52.4 & 49.5 & 21.4 & 30.7 & 25423 & 234592 & 8491 & 14797 & 6.3\\
    Tracktor (\cite{tracktor}) & Online & LA & 53.5 & 52.3 & 19.5 & 36.6 & \textbf{12201} & 248047 & 2072 & 4611 & 1.5 \\
    UMA (\cite{uma}) & Online & LA & 53.1 & 54.4 & 21.5 & 31.8 & 22893 & 239534 & 2251 & - & 5.0 \\ 
    \textbf{TADN (Ours)} & Online & TADN & \textbf{54.6} & 49.0 & 22.4 & \textbf{30.2} & 36285 & \textbf{214857} & 4869 & 7821 & 10.0 \\ 
\end{tabular*}}
\footnotetext{\footnotemark[1]Hz: Tracking inference performance. TADN performance achieved using a Nvidia Geforce GTX2080Ti. Performance may vary for different hardware configurations.}
\end{minipage}
\end{table*}

\begin{table*}
\caption{UA-DETRAC results}\label{table-resultsUADETRAC}
\centering
\begin{minipage}{\textwidth}
\resizebox{\textwidth}{!}{
\begin{tabular*}{\textheight}{@{\extracolsep{\fill}}ll*7c@{\extracolsep{\fill}}}
    \textbf{Detector + Tracker} & \textbf{DA} & \textbf{MOTA\footnotemark[1]$\uparrow$} & \textbf{MOTP\footnotemark[1]$\uparrow$} & \textbf{MT\footnotemark[1]$\uparrow$} & \textbf{ML\footnotemark[1]$\downarrow$} & \textbf{FP\footnotemark[1]$\downarrow$} & \textbf{FN\footnotemark[1]$\downarrow$} & \textbf{IDSW\footnotemark[1]$\downarrow$} \\
    \hline
    CompACT\footnotemark[2] $+$ H2T (\cite{h2t-detrac}) & UHRH & 12.4 & 35.7 & 14.8 & 19.4 & 51765 & 173899 & 852 \\
    CompACT\footnotemark[2] $+$ CMOT (\cite{cmot-detrac}) & LA & 12.6 & 36.1 & 16.1 & 18.6 & 57885 & 167110 & \textbf{285} \\
    CompACT\footnotemark[2] $+$ GOG (\cite{gog-detrac}) & minCF & 14.2 & 37.0 & 13.9 & 19.9 & 32092 & 180183 & 3334 \\
    EB\footnotemark[3] $+$ IOUT (\cite{KIOU-detrac}) & LA & 19.4 & 28.9 & 17.7 & 18.4 & 14796 & 171805 & 2311 \\
    CompACT\footnotemark[2] $+$ FAMNet (\cite{famnet}) & e2e-MDAP & 19.8 & 36.7 & 17.1 & 18.2 & 14989 & 164433 & 617 \\
    EB\footnotemark[3] $+$ DAN (\cite{deep-affinity-network}) & e2e-LA & 20.2 & 26.3 & 14.5 & 18.1 & \textbf{9747} & \textbf{135978} & 518 \\
    EB\footnotemark[3] $+$ Kalman-IOUT (\cite{KIOU-detrac}) & LA & 21.1 & 28.6 & 21.9 & 17.6 & 19046 & 159178 & 462 \\
    \textbf{EB\footnotemark[3] $+$ TADN (Ours)} & TADN & \textbf{23.7} & \textbf{83.2} & \textbf{61.2} & \textbf{8.2} & 31417 & 198714 & 2910 \\    
\end{tabular*}}
\footnotetext{\footnotemark[1]CLEAR metrics along the detector's PR-curve. Each metric value is the mean value for 10 equally spaced detection thresholds. }
\footnotetext{\footnotemark[2]\cite{CompACT-detector}, \footnotemark[3]\cite{EB-detector}}
\footnotetext{\textbf{Abbreviations} e2e: end-to-end trainable, LA: Linear Association, MDAP: Multi-Dimensional Association Problem, MHT: Multiple Hypothesis Tracking, UHRH: Undirected Hierarchical Relation Hypergraph, minCF: min cost flow network-based DA, GDA: Greedy Data Association, OSL-MN: One-Shot Learning with MatchingNet adaptation, MAIQP: Maximizing An Identity-Quantity Posterior problem}

\end{minipage}
\end{table*}

\begin{table*}
\caption{MOT17 results on private detections}\label{table-resultsMOT17private}
\centering
\begin{minipage}{\textwidth}
\resizebox{\textwidth}{!}{
\begin{tabular*}{\textheight}{@{\extracolsep{\fill}}*3l*9c@{\extracolsep{\fill}}}
    \textbf{Tracker} & \textbf{Data} & \textbf{MOTA$\uparrow$} & \textbf{IDF1$\uparrow$} & \textbf{MT$\uparrow$} & \textbf{ML$\downarrow$} & \textbf{FP$\downarrow$} & \textbf{FN$\downarrow$} & \textbf{IDSW$\downarrow$} & \textbf{Frag$\downarrow$} & \\
    \hline
    TubeTK (\cite{tubetk}) & JTA & 63.0 & 58.6 & 31.2 & 19.9 & 27060 & 177483 & 4137 & 5727 \\
    CenterTrack (\cite{centertrack}) & CH & 67.8 & 64.7 & 34.6 & 24.6 & \textbf{18498} & 160332 & 3039 & 6102 \\
    TraDeS (\cite{trades}) & CH & 69.1 & 63.9 & 36.4 & 21.5 & 20892 & 150060 & 3555 & 4833 \\
    MOTR (\cite{motr}) & CH & 73.4 & 68.6 & - & - & - & - & \textbf{2439} & - \\
    FairMOT (\cite{fairmot}) & CH+6T & 73.7 & \textbf{72.3} & 43.2 & 17.3 & 27507 & 117477 & 3303 & 8073 \\
    TrackFormer (\cite{trackformer}) & CH & 74.1 & 68.0 & \textbf{47.3} & \textbf{10.4} & 34602 & \textbf{108777} & 2829 & 4221 \\
    TransTrack (\cite{transtrack}) & CH & \textbf{74.5} & 63.9 & 46.8 & 11.3 & 28323 & 112137 & 3363 & 4872 \\
    \textbf{TADN with YOLOX (Ours)} & - \footnotemark[1]{} & 69.0 & 60.8 & 45.7 & 13.6 & 47466 & 124623 & 2955 & \textbf{4119} \\ 
\end{tabular*}}
\footnotetext{\footnotemark[1] TADN model is trained \textbf{exclusively} on MOT17 provided training data. YOLOX detector used is a community pretrained model publicly available trained on CrowdHuman (\cite{crowdhuman}), MOT17D, CityPersons (\cite{zhang2017citypersons}) and ETHZ (\cite{li2017scale}) detection datasets.}
\end{minipage}
\end{table*}

\begin{table*}
\caption{MOT20 results on private detections}\label{table-resultsMOT20}
\centering
\begin{minipage}{\textwidth}

\resizebox{\textwidth}{!}{
\begin{tabular*}{\textheight}{@{\extracolsep{\fill}}*3l*9c@{\extracolsep{\fill}}}
    \textbf{Tracker} & \textbf{Data} & \textbf{MOTA$\uparrow$} & \textbf{IDF1$\uparrow$} & \textbf{MT$\uparrow$} & \textbf{ML$\downarrow$} & \textbf{FP$\downarrow$} & \textbf{FN$\downarrow$} & \textbf{IDSW$\downarrow$} & \textbf{Frag$\downarrow$} \\
    \hline
    FairMOT (\cite{fairmot}) & CH+6T & 61.8 & 67.3 & 68.8 & 7.6 & 103440 & 88901 & 5243 & 7874 \\
    TransTrack (\cite{transtrack}) & CH & 64.5 & 59.2 & 49.1 & 13.6 & 28566 & 151377 & 3565 & - \\
    GSDT (\cite{gsdt}) & 5T & 67.1 & 67.5 & 53.1 & 13.2 & 31507 & 135395 & 3133 & 9878 \\
    TransCenter (\cite{transcenter}) & - & 67.7 & 58.7 & 66.3 & 11.1 & 56435 & 107163 & 3759 & - \\
    OUTrack-fm (\cite{outrackfm}) & CH & 68.5 & 69.4 & 57.9 & 12.2 & 37431 & 123197 & 2147 & 5683 \\
    TrackFormer (\cite{trackformer}) & CH & 68.6 & 65.7 & 53.6 & 14.6 & \textbf{20348} & 140373 & \textbf{1532} & 2474 \\
    TransMOT (\cite{transmot}) & CH & \textbf{77.4} & \textbf{75.2} & \textbf{70.1} & \textbf{9.2} & 32335 & \textbf{82867} & 1601 & - \\
    \textbf{TADN with YOLOX (Ours)} & - \footnotemark[1]{} & 68.7 & 61.0 & 57.4 & 14.3 & 27135 & 133045 & 1707 & \textbf{2321} \\ 
\end{tabular*}}
\footnotetext{\footnotemark[1] TADN model is trained \textbf{exclusively} on MOT20 provided training data. YOLOX detector used is a community pretrained model publicly available trained on CrowdHuman (\cite{crowdhuman}) and MOT20D detection datasets.}
\footnotetext{\textbf{Abbreviations} CH: CrowdHuman Dataset (\cite{crowdhuman}), JTA: Joint Track Auto dataset (\cite{jta}), 6T: 6 tracking datasets as stated in \cite{fairmot} }

\end{minipage}
\end{table*}

\textbf{MOT17 with public detections.} TADN was evaluated on the public detections track of MOT17 benchmark against similar state-of-the-art approaches (Table \ref{table-resultsMOT17}). In a public detections setting a tracker's performance can be evaluated without the variability induced by the performance of the object detection algorithm used, since all methods rely on the same detections set for training and testing. For fairness, we opted to compete against state-of-the-art online and offline methods that directly apply the tracking-by-detection paradigm and do not rely on filtering techniques to classify for the public detections track while using private detections for inference. Such techniques apply for trackers of the joint detection and tracking scheme via either imposing overlap constraints for target initialization based on the public detections, such as in the work of \cite{bytetrack}, or via limiting the maximum number of birth candidates at each frame to the number of public detections available (\cite{transcenter}). Furthermore, to achieve high performance compared to state-of-the-art, many methods use external datasets during training additionally to MOT17 train set, such as CrowdHuman (\cite{crowdhuman}) in TrackFormer(\cite{trackformer}) or HiEve (\cite{hieve}) in TransMOT(\cite{transmot}). Our method, on the other hand, is trained solely on the MOT17 provided training data. 

To summarize, for a fair comparison, we opted to compare against proven and well established methods that (a) compete in the public detections track following a tracking-by-detection paradigm, (b) use exclusively the data provided by the benchmark and (c) represent a variety of different data association methods. Methods considered are officially published and peer-reviewed in the MOTChallenge benchmark. Quantitative evaluation is based on CLEAR metrics (\cite{clear-metrics}) which provide a global tracking performance measures such as MOTA and IDF1, as well as a series of more granular metrics regarding important aspects of tracking procedures, such as fragmentation of trajectories (Frag) and identity switches of targets (IDSW). 

TADN outperformed all comparable online methods regarding MOTA, Mostly Lost (ML), False Negative (FN) metrics while yielding highly competitive performance for IDF1 and Mostly Tracked (MT). Similarly, compared to offline methods TADN offers comparable performance despite the by-design advantage of these methods to recreate a target's trajectory using both past and future observations. The results of our simple tracking algorithm prove that TADN can provide a good associative performance for matching targets to detections that can substitute more traditional data association techniques. On the other hand, our method was more sensitive to False Positives (FP), ID Switches (IDSW) and Frag metrics. High FP values denote a frequent drift of a target relative to the ground-truth. This behaviour is mostly due to the simplistic motion model used that fails to predict accurate future locations for the targets. That also leads in a rather straightforward manner into high Frag values. This phenomenon is also enlarged due to the design of ``set-inactive" threshold $T_h$ where targets with few detections can more easily be set to inactive leading to a frequent spawn of new trajectories. Constantly spawning new targets leads to high IDSW values since each ground-truth trajectory is assigned to a series of smaller tracks created by our method. Furthermore, the fact that we did not employ any re-identification module for target ``re-birth" or occlusion handling techniques along with the frame-by-frame training strategy formulation have a negative impact on those aforementioned metrics.

\textbf{MOT17 and MOT20 with private detections}. For completeness we tested also TADN coupled with a private detector to compete in the private detection track of MOT20 and MOT17. In either case we used a community pretrained YOLOX (\cite{yolox}) object detector for each benchmark separately, while we limited the training of TADN model exclusively on the provided training tracking data. As implied before, under such a context we cannot draw definitive conclusions regarding the associative performance of TADN, but we can obtain a good estimate of the potential of our method for the generic MOT problem. For both of these benchmarks we pit against several state-of-the-art algorithms without being limited to benchmark's training data exclusively or restricted to purely tracking-by-detection methods. 

Regarding MOT17 (Table \ref{table-resultsMOT17private}), TADN performs adequately well for the MOTA, IDSW, Frag, MT and ML metrics. Compared to other methods TADN outperforms several methods for the MOTA metric such as CenterTrack (\cite{centertrack}), TraDes (\cite{trades} or TubeTK (\cite{tubetk}). In any case, TADN achieves such performance without needing extra tracking training data or employing sophisticated solutions for appearance and motion modeling. It should be noted however that in that case tracking performance is heavily related to the detection performance of the private object detector. Strong variability should be expected if another detector is used, however that is the case for every method we compare to. As a final remark, we observe some similarities compared to MOT17-public regarding the behaviour of certain CLEAR metrics, especially the subpar performance of FP metric. 

Regarding MOT20 (Table \ref{table-resultsMOT20}), we observe a different behaviour for the performance of TADN. In that case, while maintaining a competitive MOTA performance, TADN performs best on Frag, IDSW and FP metrics while IDF1 and ML metrics suffer the most. We attribute that behaviour to two main reasons; firstly the scenes in MOT20 differ vastly to those of MOT17 regarding simultaneous count (heavily crowded) and motion dynamics of targets. Secondly, the performance of the private detector used varies between the two benchmarks. As a general remark, TADN seems to perform better overall than methods that use extra tracking training data such as GSDT (\cite{gsdt}), FairMOT (\cite{fairmot}) or TrackFormer (\cite{trackformer}).

As a final note, the results attained provide some definitive clues that TADN could potentially be a good alternative for data association if incorporated into a more complex MOT model architecture either under a tracking-by-detection scheme or integrated into a joint detection and tracking model. We intend to explore this possibility in a future work.

\textbf{UA-DETRAC}. In order to study whether our method could generalize into a slightly different tracking benchmark, we tested TADN's performance on the UA-DETRAC dataset (Table \ref{table-resultsUADETRAC}). Using the EB (\cite{EB-detector}) detector, our proposed method outperformed competing algorithms against several metrics, i.e. PR-MOTA, PR-MOTP, PR-MT and PR-ML among a variety of selected trackers. Since these metrics are highly dependent on the detector's performance, in Figure \ref{fig_pr_mota} we observed that for a suitable detection threshold, our novel approach achieved over 35\% MOTA. High PR-MOTP values are to be expected from our method due to its trend to create several mini trajectories that span across large segments of the ground-truth trajectories. This can also be observed in the low performance on the PR-FP, PR-FN and PR-IDSW metrics that are heavily correlated to the creation of highly fragmented trajectories since our method was focused on solving a frame-by-frame association problem where no specialized measures for long-term associations are used such as occlusion handling, advanced motion model or re-identification modules, as has also been previously discussed for MOT17. 

\textbf{Comparison to other DA methods}. Methods that solve a linear assignment problem explicitly which are either end-to-end trainable (\cite{famnet}, \cite{how-to-train}) or not (\cite{tracktor}, \cite{dman}) achieve relatively high MOTA values in the cost of inference speed performance. MHT trackers  \cite{mht-2} tend to generate less fragmented trajectories with high MOTA and IDF1 values, however they add extra computational cost during inference. TADN's inference speed, including both feature extraction and association, as shown in Table \ref{table-resultsMOT17}, was significantly higher compared to all other methods and actually only bound to the appearance model inference performance since TADN has a much smaller computational overhead, yielding about 10 Hz for the association part. This renders our DA alternative methodology suitable for deployment even on embedded devices to provide real-time tracking.

\subsection{Qualitative Evaluation on MOT17 and UA-DETRAC}

\begin{figure*}[h]
  \includegraphics[width=\textwidth]{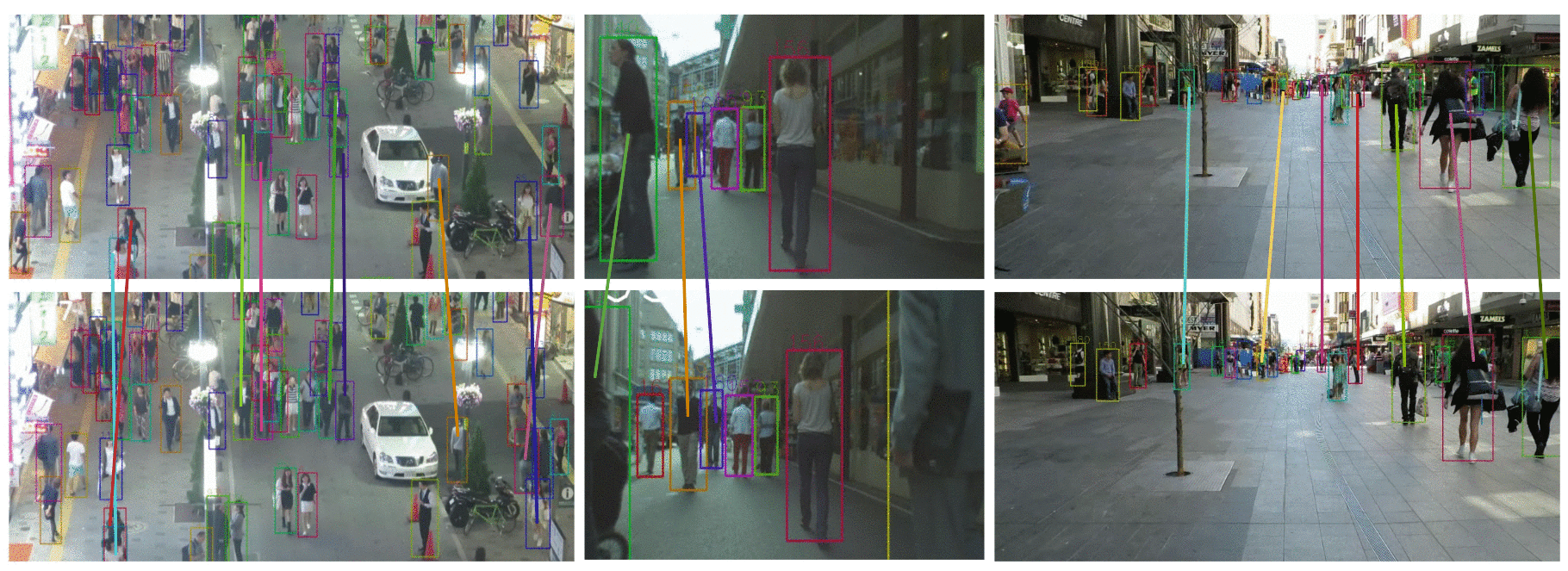}
  \caption{Success cases from MOT17 test set TADN results in presence of partial occlusion, various scene geometries and illumination conditions. Each pair is $\sim$ 30 frames apart.}
  \label{fig-success}
\end{figure*}

We showcase in Figure \ref{fig-success} that in most cases, regarding the MOT17 dataset, our method successfully reconstructs ground truth trajectories. The associative performance of our method seems to be robust even in highly crowded scenes with different camera configurations regarding resolution, scene geometry or illumination conditions. Robustness in target scale changes and partial occlusions demonstrate our model's capacity to associate positional and appearance features between detections and existing trajectories in the scene. Similar behavior was observed on UA-DETRAC qualitative results as shown on the left pair of images in Figure \ref{fig-detrac}.

\begin{figure}[H]
\centering
\begin{tikzpicture}
\begin{axis}[%
    height=4cm,
    width=\columnwidth,
    legend style={at={(0.4,0.25)},anchor=west},
    x label style={at={(axis description cs:0.5,0.1)},anchor=north},
    y label style={at={(axis description cs:0.03,0.5)},anchor=north},
    xlabel={Detection threshold},
    ylabel={MOTA \%},
    xmin=0, xmax=1.0,
    ymin=0, ymax=40,
    xtick={0,0.2,0.4,0.6,0.8,1.0},
    ytick={0,10,20,30,40},
    ymajorgrids=true,
    grid style=dashed,
    scatter/classes={%
    a={mark=o,draw=black}}
]
    
\addplot[
    color=black,
    mark=square,
    ]
coordinates {
(0,0)
(0.1,7.7)
(0.2,13.2)
(0.3,17.1)
(0.4,20.7)
(0.5,24)
(0.6,27.6)
(0.7,31.2)
(0.8,34.4)
(0.9,37.2)
};
\legend{EB + TADN}

\end{axis}
\end{tikzpicture}
\caption{MOTA on UA-DETRAC test set for detection thresholds along the PR-curve.}
\label{fig_pr_mota}
\end{figure}

\begin{figure*}[!ht]
  \includegraphics[width=\textwidth]{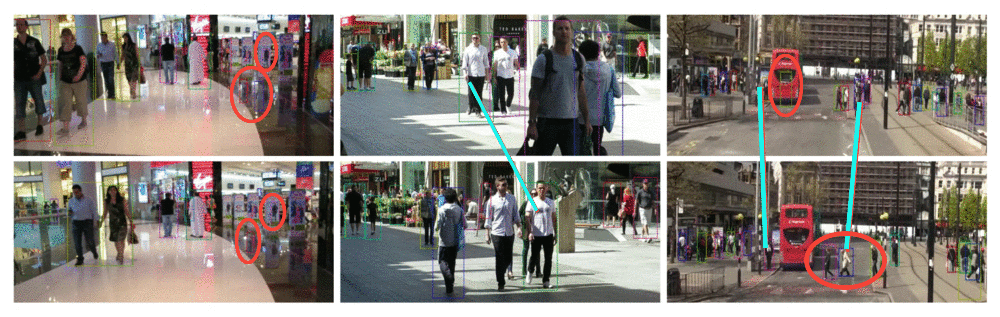}
  \caption{Failure examples from MOT17 test set TADN results in presence of poor detections and strong occlusions. Cyan lines: id-switches. Red ellipses: Falsely tracked objects.}
  \label{fig-failure}
\end{figure*}

\begin{figure*}[ht]
  \includegraphics[width=\textwidth]{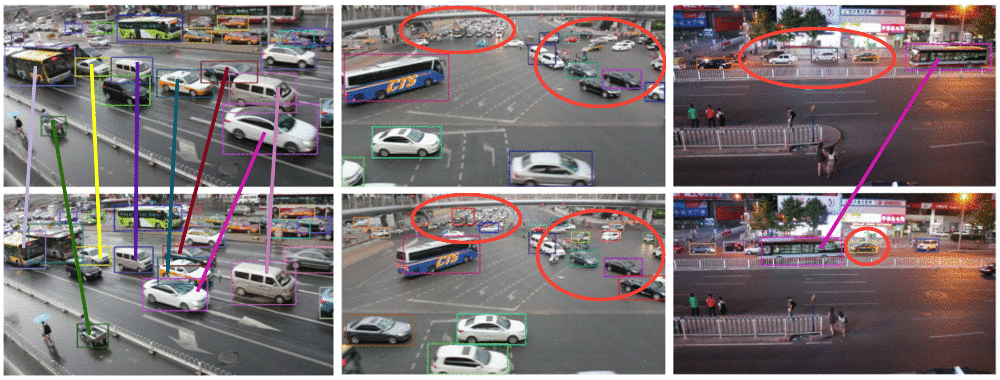}
  \caption{UA-DETRAC test set TADN results. Each pair is $\sim$ 40 frames apart. Left: Success cases with partial occlusions. Mid, Right: Failure cases due to large occlusions, diverse motion dynamics and lack of detections depicted by red colored ellipses.}
  \label{fig-detrac}
\end{figure*}

On the other hand, mostly due to the employed rather simple MOT framework lacking any occlusion handling or re-identification modules produced some failure cases, as shown in Figure \ref{fig-failure}. In detail, there were some identity switch occurrences where two targets in close proximity had a very similar appearance like the middle example of Figure \ref{fig-failure}. Also, some tracking errors correlated mostly with false detections provided by the MOT17 benchmark such as those on the first example in Figure \ref{fig-failure} where some pedestrians' reflectances were detected consistently and thus tracked by our algorithm. Despite using CMC features, sequences characterized by a moving camera in general tended to perform worse than those with a static camera configuration. Furthermore, our motion model failed to reliably predict future target locations in cases where targets move in complex paths on a crowded scene generating a lot of full occlusions.

Similar problems were also highlighted on UA-DETRAC qualitative results in Figure \ref{fig-detrac}. Here, there were more straightforward cases of total occlusions and highly different motion dynamics for each vehicle in the scene. In either case, our method consistently failed to sustain long-term trajectories and many identity switches occur. Overall, for both datasets, we observed that most errors can be traced back to either false or missing detections or to the simplicity of specific submodules for our MOT approach such as the motion and appearance models used. However, in this work we focused on the frame-by-frame associative performance of our model that is only indirectly linked to such cases.

\subsection{Ablation study}

To achieve optimal performance, our tracker and TADN in particular require finetuning of multiple hyperparameters and examination of several configuration options. To this end, we conducted a thorough two-part ablation study where, firstly, we investigated multiple architecture configurations for the Transformer architecture in TADN module, while in the second part we explored several other options related to our tracking framework. All ablation were conducted on a 50/50 per sequence split of the original MOT17 training data due to limitations of the test server. In  order to assess the results, we used 4 crucial CLEAR metrics, namely MOTA, IDSW, FP and FN.

\textbf{TADN architecture}. We conducted a series of experiments altering three principal Transformer parameters, i.e. number of heads for the multi-head attention layers, number of encoder layers and number of decoder layers (Table \ref{Table_Ablation1}). In all experiments, it is assumed a dual-branch TADN module using both appearance and positional features coupled with a Kalman motion model with ECC-based CMC and the $ulbr_1$ metric as the groundtruth assignment metric during training. All configurations achieved a MOTA greater than 60\%, while the most performant combination was achieved by the simplest Transformer architecture yielding also the less ID switches, FP and FN. More complicated models seemed to perform worse, however this might be due to insufficient data that led our model to underfit.

\begin{table*}
\caption{TADN architecture experiments}
\begin{center}
\begin{minipage}{\textwidth}
\begin{tabular*}{\textwidth}{@{\extracolsep{\fill}}*7c@{\extracolsep{\fill}}}
    \multicolumn{3}{c}{\textbf{Architecture parameters}}& \multicolumn{4}{c}{\textbf{Metrics}}\\

    \textbf{Heads} & \textbf{Encoder Layers} & \textbf{Decoder Layers} & \textbf{MOTA$\uparrow$} & \textbf{IDSW$\downarrow$} & \textbf{FP$\downarrow$} & \textbf{FN$\downarrow$} \\
    \hline
        \textbf{2} & \textbf{2} & \textbf{2} & \textbf{62.2} & \textbf{454} & \textbf{2021} & \textbf{17969} \\
        4 & 2 & 2 & 61.5 & 472 & 2060 & 18273 \\
        2 & 4 & 4 & 60.6 & 598 & 2485 & 18221 \\
        4 & 4 & 4 & 61.0 & 605 & 2545 & 18223 \\
        2 & 2 & 4 & 60.8 & 522 & 2383 & 18296 \\
    \end{tabular*}
    \end{minipage}
    \end{center}
    \label{Table_Ablation1}
\end{table*}

\begin{table*}
    \caption{Ablation on MOT configuration options}
    \begin{center}
    \begin{minipage}{\textwidth}
    \begin{tabular*}{\textwidth}{@{\extracolsep{\fill}}*5c|*4c@{\extracolsep{\fill}}}
    \multicolumn{5}{c}{\textbf{Configuration options}} & \multicolumn{4}{c}{\textbf{Metrics}}\\
    \textbf{B} & \textbf{F} & \textbf{M.M.} & \textbf{CMC} & \multicolumn{1}{c}{\textbf{M}} & \textbf{MOTA$\uparrow$} & \textbf{IDSW$\downarrow$} & \textbf{FP$\downarrow$} & \textbf{FN$\downarrow$} \\
    \hline
        D & A,P & Kalman & $\checkmark$ & ulbr1 & \textbf{62.2} & 454 & 2021 & \textbf{17969} \\
        D & A,P & Linear & - & ulbr1 & 60.2 & 642 & 2140 & 18683 \\
        D & A,P & Kalman & - & ulbr1 &60.7 & \textbf{450} & 2070 & 18716 \\
        \hline
        S & A,P & Kalman & $\checkmark$ & ulbr1 & 61.5 & 460 & 2075 & 18289 \\
        \hline
        D & A,P & Kalman & $\checkmark$ & IoU & 60.8 & 425 & 2076 & 18677 \\
        \hline
        D & P & Kalman & $\checkmark$ & ulbr1 & 47.5 & 538 & \textbf{1603} & 27412 \\
        D & A & Kalman & $\checkmark$ & ulbr1 & 61.3 & 500 & 2103 & 18276 \\
        \hline \hline
        \multicolumn{2}{c}{\textbf{T.S.A.}$^*$} & Kalman & $\checkmark$ & ulbr1 & 62.9 & 702 & 1003 & 18358 \\
    \end{tabular*}
    \footnotetext{$^*$ Training Strategy-only Assignments. Refers to an ideal tracker where the assignments are directly derived from ground-truth data via our proposed training strategy.}
    \footnotetext{\textbf{Abbreviations:} \textbf{B}: \# Branches, \textbf{F}: Selected features, \textbf{M.M.}: Motion model, \textbf{CMC}: Camera Motion Compensation, \textbf{M}: Assignment metric, D: Dual branch TADN, S: Single branch TADN, A: Appearance features, P: Positional features}
    \end{minipage}
    \end{center}
    \label{Table_Ablation2}
\end{table*}

\textbf{MOT configuration options}. A series of experiments with different configurations was conducted examining each time (i) motion model selection and CMC benefits (ii) single or dual branch architectures, (iii) assignment metric during training and (iv) selected features fed to the TADN module (Table \ref{Table_Ablation2}). Overall, our best model proved to be the one with dual branch TADN module using both appearance and positional features coupled with a CMC-enabled Kalman motion model and $ulbr_1$ as the assignment metric during training yielding the best MOTA and FN values and near-best IDSW and FP values. This configuration was considered thus the baseline for all experiments.

By employing a linear motion model the overall performance dropped significantly for every metric in comparison to a Kalman motion model. Enabling CMC further reduced FN metric and boosted MOTA metric by 1.5\%. Regarding the single branch configuration, it yielded similar performance for IDSW and FP metrics, however it had lower MOTA value by 0.7\% and induced more false negatives showing that the dual branch configuration could reconstruct successfully a bigger part of the ground truth trajectories by a significant margin. 

Using IoU as the assignment metric during training seemed to perform significantly worse than the proposed $ulbr_1$ metric in all considered metrics. This fact indicates that $ulbr_1$ is more suitable for training such models since it performs better in cases we have minimal or no overlap between ground truth and predicted bounding boxes during training time. Last, in order to measure the importance of either appearance or positional features that are fed to TADN, we observed that by employing only positional features MOTA dropped by 14.7\%. However, that configuration yielded the less false positives among all configurations denoting that positional features are very important to generate as less as possible fragmented trajectories. On the other hand, using only appearance features the performance was close to the baseline (lacking a 0,9\%). However, this configuration produced more identity switches compared to the one using both appearance and positional features. This was expected since in general two targets may have similar appearance features but can vary significantly in their location in the scene.

The last row in Table \ref{Table_Ablation2} refers to a hypothetical tracker where in each step the final assignments were not derived by our model, but from $LAM$ using our training strategy. This experiment was indicative of the expected tracking performance especially during the early stages of training. This measurement directly influenced the final trained model but also gave us an estimate of the maximal performance to be expected given this training strategy. We observed that our model reached this theoretical performance for MOTA metric, outperformed it for IDSW and FN, while underperformed for FP.

\begin{figure*}[ht]
\centering
 \begin{tikzpicture}
    \begin{axis}[
      height=2.2cm,
      width=10cm,
      scale only axis,
      x label style={at={(axis description cs:0.5,0.1)},anchor=north},
      xmin=0.1,xmax=0.8,
      ymin=60, ymax=65,
      xlabel={$IoU\; T_{tgt2det}$ threshold},
      xtick={.15,.3,.45,.6,.75},
      ymajorgrids=true,
      grid style=dashed,
      ylabel={MOTA \%},
      axis y line*=left,
      axis x line*=bottom]
        
        \addplot[color=black,mark=*,mark options=solid]
        coordinates {
        (0.15,62.6)
        (0.3,63.5)
        (0.45,63.5)
        (0.6,62.5)
        (0.75,56.0)
        };\label{iou}
        
    \end{axis}
    \begin{axis}[
      height=2.2cm,
      width=10cm,
      scale only axis,
      x label style={at={(axis description cs:0.5,1.65)},anchor=north},
      xmin=-0.22,xmax=-0.08,
      ymin=60, ymax=65,
      xtick={-0.21,-0.17,-0.13,-0.09},
      ytick={0},
      xlabel={$ulbr_1 \; T_{tgt2det}$ threshold},
      axis y line*=right,
      axis x line*=top]
      \addplot[dashed,color=black,mark=triangle,mark options=solid]
        coordinates {
        (-0.09,61.7)
        (-0.11,62.3)
        (-0.13,62.9)
        (-0.15,62.9)
        (-0.17,63.2)
        (-0.19,62.9)
        (-0.21,62.5)
        };
        \legend{$ulbr_1$}
        \addlegendimage{/pgfplots/refstyle=iou}\addlegendentry{$IoU$}
    \end{axis}
  \end{tikzpicture}
\caption{MOTA values for different $IoU$ and $ulbr_1$ thresholds. $IoU$ achieves better MOTA overall, but is more sensitive to small perturbations.}
\label{fig_tgt2det}
\end{figure*}
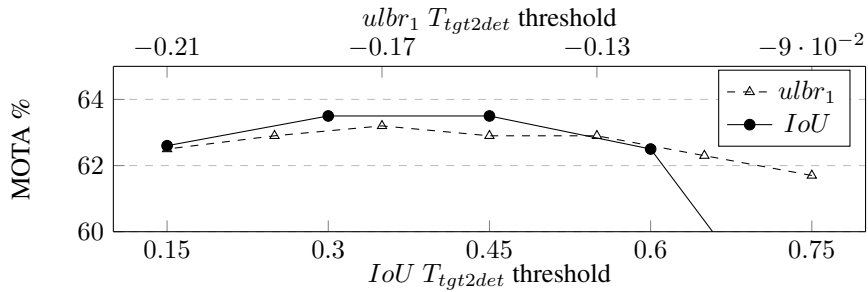

Last, we conducted a series of sensitivity analysis experiments for different values of $T_{tgt2det}$ for both assignment metrics $ulbr_1$ and $IoU$ where we used the aforementioned hypothetical tracker (Figure \ref{fig_tgt2det}). Comparing these two metrics, while $IoU$ was able to achieve a higher overall MOTA score, it was very sensitive to the threshold value selection. In contrast, the proposed $ulbr_1$ consistently yielded MOTA scores greater than 62.5\% for a significant interval of possible thresholds. All in all, for the experiments conducted using the $IoU$ and $ulbr_1$ metric we set $T_{tgt2det}=0.3$ and $T_{tgt2det}=-0.13$ respectively that despite not being the one yielding the best MOTA overall, it was the best compromise between attaining a high MOTA score and being restrictive enough regarding box overlaps. 

\section{Conclusions}\label{sec5}
Solving a linear data association problem adds significant complexity and in most cases can introduce various problems to multiple object tracking. In this work, we introduced a novel data association method for multiple object tracking via formulating an assignment decision network based on a Transformer architecture. Using our end-to-end training strategy we achieved highly competitive performance compared to popular trackers that employ state-of-the-art data association techniques on challenging benchmarks without using any sophisticated submodules refering to appearance or motion modeling. Regarding deployment scenarios, TADN could be a viable alternative in several application fields, especially in embedded environments where real-time performance is crucial. In short, TADN could be a good option for vehicle or pedestrial monitoring, realtime transportation analytics or similar applications, while for others FP-critical applications, such as medical, TADN should be deployed on a case-by-case basis. For optimal results pairing TADN with a suitable object detector is recommended, e.g. using a realtime capable detector for embedded applications or a high performant one, like YOLOX, for FP-sensitive ones.

Overall, TADN demonstrated significant potential in tracking applications, however based on our study it was limited by two factors; our rather simple tracking-by detection framework employed and our training strategy that derives target associations considering exclusively two consecutive frames. Since during training time ground truth trajectories are available for the whole sequence length, the training strategy could significantly benefit from computing assignments in a non causal way and potentially yielding better performance especially with metrics related to long-term association performance such as IDSW or Frag. Such a modification does not compromise the online characterization of our approach since it only affects processes during training time, while during testing causality is maintained. Also, using an external detection module has a significant benefit of decoupling the detection from the assignment pipeline hence rendering our approach transferable to new application domains. To improve performance in future work, many state-of-the-art sophisticated modules can be directly incorporated into our tracking pipeline such as occlusion handling, better appearance and motion modeling, more accurate detections and re-identification module.

\section*{Acknowledgments}\label{sec6}
This work was partly funded by the BiCubes research project and the scholarship of the Research Commitee of the National Technical University of Athens.

\bibliographystyle{bibstyles}
\bibliography{references}

\end{document}